\documentclass[letterpaper, 10 pt, conference]{ieeeconf}  

\IEEEoverridecommandlockouts                              

\usepackage{amsmath,amssymb,bm}
\usepackage[Symbol]{upgreek}
\usepackage{graphicx}
\usepackage{graphics}
\usepackage[tight]{subfigure}
\usepackage{floatrow} 
\usepackage{multirow}
\usepackage{array}
\usepackage{enumerate}
\usepackage{sidecap}
\usepackage[all]{xy}
\usepackage{authblk}
\usepackage{comment}
\usepackage{changepage}   
\usepackage{wrapfig}
\usepackage{url}
\usepackage{tabularx,pbox,booktabs,caption}
\usepackage{flushend}
\usepackage{anyfontsize}
\usepackage{float}
\usepackage{todonotes}
\usepackage[percent]{overpic}
\graphicspath{{./figs/}}
\usepackage{amssymb}
\usepackage{amsfonts}

\usepackage{color}
\newcommand{\LL}[1]{\textcolor{red}{{\bf LL:} #1}}
\newcommand{\ZC}[1]{\textcolor{orange}{{\bf ZC:} #1}}

\title{\large \bf Creating Navigable Space from Sparse Noisy Map Points}

\author{Zheng Chen and Lantao Liu
\thanks{\newline Zheng Chen and Lantao Liu are with the School of Informatics, Computing, and Engineering  at Indiana University, Bloomington, IN 47408, USA. E-mail:
        {\tt\small \{zc11, lantao\}@iu.edu} 
}
}

\begin{document}

\maketitle
\thispagestyle{empty}
\pagestyle{empty}

\begin{abstract}
To facilitate robot navigation, we present a framework for creating navigable space from sparse and noisy map points generated by SLAM methods with noisy sensors or poor features. 
Our method incrementally seeds and creates local convex regions free of obstacle points along robot's trajectory.
Then a dense version of point cloud is reconstructed through a map point regulation process where the original noisy map points are first projected onto a series of local convex hull surfaces, 
after which those points falling inside the convex hulls are culled. 
The regulated and refined map points 
 will not only ease robot navigation and planning, but also allow human users to quickly recognize and comprehend the environmental information.
We have validated our proposed framework using both a public dataset and a real environmental structure, and our results reveal that the reconstructed navigable free space has small volume loss (error) comparing with the ground truth, and the method is highly efficient, 
allowing real-time computation and online planning.
\end{abstract}


\section{Introduction}

Simultaneous localization and mapping (SLAM) has been shown as an excellent solution for autonomous robots  in GPS-denied environments~\cite{thrun2005probabilistic}.
However, the map building process typically depends on some prescribed path following which the environmental features can be collected.   
Planning motion within the built map is a well-known challenging task~\cite{kim2013perception}, especially when the map quality is low due to noisy sensors (e.g., with a low-end LiDAR or sonar system) or noisy features (e.g., with visual perception).


We are interested in the problem of reconstructing navigable space from a cloud of 3D sparse and noisy map points that are produced from existing odometry or mapping methods~\cite{mur2015orb,forster2014svo}. 
Our work does not improve the SLAM or feature representation/processing techniques,
instead, 
the objective is to {\em conservatively} extract navigable space by {\em regulating} given sparse and noisy map points, the result of which can then be used for future collision-free navigation and motion planning.

To maximize the constructed free space from the noisy and irregular map points, 
we opt to use a convex region growing strategy~\cite{deits2015computing} 
which iteratively looks for a sequence of polyhedra with maximum inscribed ellipsoids using quadratic programming and semidefinite programming. 
We then develop a framework by incrementally seeding and creating local convex regions along a robot's trajectory.
To build a complete navigable volume (tunnel) for the entire map, 
we reconstruct a regulated version of point cloud which looks similar to the outputs of high-definition ranging/depth sensors.
The regulation is built on a projection process on the generated convex hulls. 
This is done by projecting the original noisy points around the polyhedra onto the surfaces of those polyhedra, after which those points falling inside the convex hulls are culled. 
The regulated point cloud also allows human users to quickly recognize and abstract the environmental information.


\begin{figure} [t] 
  \centering
  \subfigure[]
  	{\label{fig:intro_1}\includegraphics[width=1.6in]{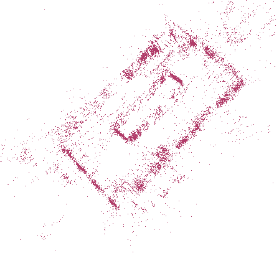}}
  	\ \ 
  \subfigure[]
  	{\label{fig:intro_2}\includegraphics[width=1.6in]{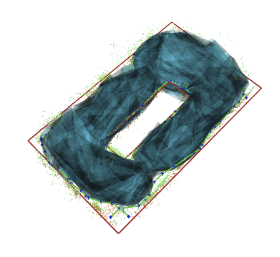} \vspace{-15pt}}
  	\ \ 
  \subfigure[]
  	{\label{fig:intro_3}\includegraphics[width=1.6in]{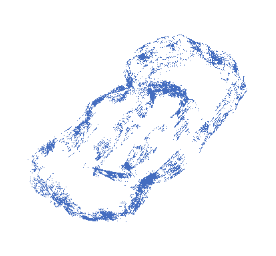}}
  	\ \ 
  \subfigure[]
  	{\label{fig:intro_4}\includegraphics[width=1.6in]{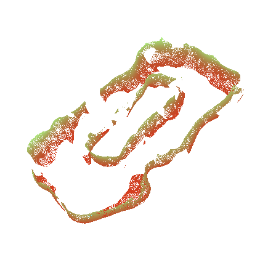} \vspace{-15pt}}
  \caption{\small An illustration of our proposed method applied to a real environment. (a) The original map point cloud generated by a sparse visual SLAM method. (b) After filtering out outlier (distant) points, the free space is computed. The free space consists of a series of overlapping convex hulls which capture the shape of the free space while excluding any points inside. (c) 
  For better visualization and environmental abstraction, we regulate the original noisy map points by projecting them onto the facets of the convex hulls. (d) 
  The projected point cloud is further refined to be denser and smoother. 
  }
\label{fig:intro}  \vspace{-10pt}
\end{figure}

To the best of our knowledge, this is the first time to extract free space  from 3D sparse and noisy map points 
through integrating point regulation and space convexification operations.




\section{Related Work}

Point cloud processing methods have been well studied in past years. 
Existing point cloud methods can be categorized as surface reconstruction, volume reconstruction, model fitting, and kernel-based regression frameworks.
Specifically, a large set of early work aims to build surfaces from point clouds. 
For instance, {moving least squares}~(MLS) based methods~\cite{scheidegger2005triangulating,fleishman2005robust} were developed to reconstruct point clouds output from laser scans; projection-based greedy strategies~\cite{marton2009fast} were used to achieve incremental surface growing and building.
Signed distance function~\cite{hoppesurface} and Bayesian method~\cite{diebel2006bayesian} have also been investigated for surface reconstruction.
Some other works, like \cite{lovi2010incremental},~\cite{romanoni2015incremental}, and \cite{ling2017building}, adopted volume carving mechanisms to obtain free space given a set of points. Typically these methods first decompose the space into cells using 3D triangulation techniques, and then the visibility constraints are used to carve out those cells passed by the visibility line. 
Different from the surface-based and volume-based reconstruction schemes,  RANSAC based model fitting methods~\cite{nan2017polyfit,isack2012energy} have been used to capture the spatial structure of the given set of points. 
Online kernel-based learning methods~\cite{hadsell2009accurate} have also been proposed to implement terrain estimation from the point cloud of a LiDAR scanner.

Related work also includes various  mapping approaches as our work relies on the 3D map points generated from existing mapping methods. 
Existing map forms for robots navigation include, e.g., occupancy grid map~\cite{elfes1989using}, 3D OctoMap~\cite{hornung2013octomap}, signed distance map~\cite{oleynikova2016signed,oleynikova2016voxblox}, topological map~\cite{blochliger2018topomap,oleynikova2018sparse}, and convex region growing map~\cite{deits2015computing,deits2015efficient}, etc. 

However, in the problem where only sparse and noisy map points are provided as inputs, existing point cloud processing methods expose limitations. 
First, the majority of methods~\cite{scheidegger2005triangulating,fleishman2005robust,marton2009fast,hoppesurface} assume that the points are captured by ranging sensors such as high-definition LiDARs, sonars or depth cameras, and therefore the points are dense and evenly distributed like a mesh surface. 
We cannot use those approaches, because we are not able to get  well estimated normal and curvature of point set from those techniques. 
Volume reconstruction methods~\cite{lovi2010incremental, romanoni2015incremental,ling2017building} need to implement 3D triangulation for all the points. This however is not necessary if building a navigable space is the final goal, i.e., we want to build a map with only a minimal set of points defining the free space instead of using all the (possibly noisy) points. In addition, the computation requirement for the 3D triangulation and the post-processing can be prohibitive if all points are triangulated. Plane fitting methods~\cite{nan2017polyfit, isack2012energy} usually fail too in our case due to the high ambiguity of point  structures.

\begin{figure} 
  \centering
      \includegraphics[width=3.2in]{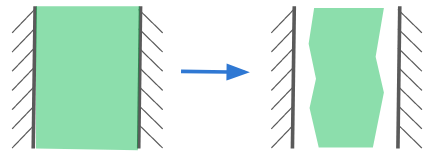}
  \caption{\small Each black solid line represents an obstacle boundary. The green region represents free space. The left figure shows an ideal situation. However, if constrained by noisy obstacle points, certain space may be lost as shown on the right.
  }
\label{fig:free_allow} \vspace{-10pt}
\end{figure}

\section{Method Overview}
\label{sect:method}


An autonomous robot needs to know clearly the obstacle space and free space before taking actions.
We desire the map that describes the environment to include precise 3D information of obstacles. 
However, this is challenging when the environment is represented as a set of sparse and noisy map points. 
One possible way to ``relax" this problem is to build an approximate navigable space with conservative surfaces to ``bound" (represent) obstacles. 
In other words, we aim at obtaining an approximation of {\em free} and {\em safe}, navigable space instead of an exact representation of the obstacle surfaces. See Fig.~\ref{fig:free_allow} for an illustration. 
The input of our work is the 3D map points (point cloud) generated from existing range-based or vision-based SLAM methods\footnote{Arguably, there are scenarios that few or no features can be detected so few or no map points can be generated. Feature detection and recognition is beyond the scope of this work. We assume that a minimal set of features can be detected so that SLAM methods, such as the visual SLAM illustrated in Fig.~\ref{fig:chara}, can at least proceed and a map point cloud can be produced.}.

\subsection{Challenges}

\begin{figure} \vspace{-5pt} 
  \centering
       \includegraphics[width=3.7in]{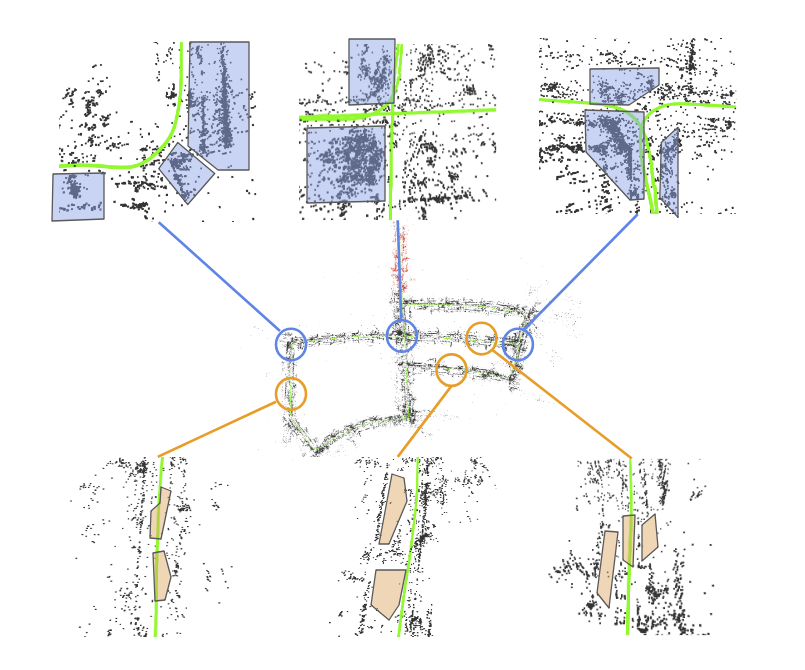}
  \caption{\small Illustration for characteristics of the map points generated from a sparse visual SLAM. The middle row is the original result from KITTI sequence 05. The blue boxes in the top row show high density areas in which the local density of points is high, but the points are not distributed like a mesh/surface (due to the estimation errors of visual SLAM, they are distributed in 3D volumes). The orange boxes in the bottom row show the zero density areas where there is no point inside the boxes, causing discontinuity. 
  }
\label{fig:chara} \vspace{-10pt}
\end{figure}

If the ranging sensors are noisy or environmental features are barren and poor, 
the produced map points can be very irregular, resulting in great difficulty for describing 3D objects. 
Fig.~\ref{fig:chara} demonstrates a mapping result with visual SLAM.
Specifically, there are a couple of challenges: 
\begin{itemize}
   \item
   {\em Large noise on point position}. Ideally, the map points of a surface (e.g., a wall)  should form a plane, but in reality the obtained map points are rather scattered randomly in 3D around the surface area, leading to great difficulty for reconstructing a surface. 
   \item 
   {\em Misleading information}. Different from conventional data processing methods where a small number of outliers are viewed as noise points, in our context the outliers could take a big fraction and cannot be simply discarded before points are regulated.
  \item
   {\em High variation on the point density}. 
   If the points are converted from visual features, 
   oftentimes the converted 3D points are not evenly distributed in the field of view, regardless of the feature detectors/descriptors employed. 
   See Fig.~\ref{fig:chara}.
\end{itemize}

Typical existing surface and volume reconstruction methods (e.g., ~\cite{fleishman2005robust,marton2009fast,hoppesurface,ling2017building}) do not work 
as they usually focus on triangulation for all the points but do not provide estimation of normal and curvature information from sparse and noisy points.
\subsection{Free Space Extraction and Construction}

Instead of reconstructing the point cloud (obstacles) as in most existing work~\cite{fleishman2005robust,marton2009fast,nan2017polyfit},
we are interested in building the free space directly through discarding the ``obstacle space".
An additional advantage of doing this is that we immediately obtain the obstacle-free area in which we can model  planning modules directly. 

Specifically, 
we opt to grow a convex hull under the spatial constraints of the map points. We adopt to use the Iterative Regional Inflation by Semidefinite (IRIS) method~\cite{deits2015computing}. 
IRIS is a convex optimization process to find a large and convex region given a set of constraints in 3D space.
The method alternates between searching for a set of hyperplanes that separate a convex space from the obstacles using quadratic programming and searching for the maximum ellipsoid which inscribes the convex space previously found using semidefinite programming.
 
Formally, an ellipsoid is represented as an image of a unit ball:
\begin{equation}
    \mathcal{E}=\left \{x = \textbf{C}\tilde{x} + \textbf{d} \, | \left \| \tilde{x} \right \|_{2} \leq 1\right \}
\end{equation}
where $x$ is the 3D points, $\tilde{x}$ represents the points inside a unit ball. \textbf{C} and \textbf{d} are the coefficients of the affine transformation.

The polyhedron (3D polytope) is represented by a set of linear inequalities:
\begin{equation}
    \mathcal{Q} = \left \{ x\, | \textbf{A}x\leq \textbf{b} \right \}
\end{equation}
where $\textbf{A}x\leq \textbf{b}$ denotes a series of constraints and $\textbf{A}, \textbf{b}$ are a coefficient matrix and a vector, respectively. 

The notation of the ellipsoid takes a form of affine transformation, which has a straightforward meaning when we try to maximize the volume of the ellipsoid: the volume is proportional to the determinant det$\textbf{C}$. Let $v_k$ be the vertices of the obstacles and $\mathcal{\zeta}$ be the set of obstacles. The problem could be formulated as follows:
\begin{equation}\label{eq_opt1}
\begin{aligned}
& \underset{\textbf{A, b, C, d}}{\text{maximize}}
& & \text{log\,det} \textbf{C} \\
& \text{subject to}
& & a^T_j v_k \geq b_j,\; \text{ for all points}\;  v_k\in \zeta_j,\\
&&& \hspace{1.91cm}\text{for}\; j = 1,\cdots, N \\
&&& \hspace{1.91cm}\text{for}\; k = 1,\cdots, K \\
&&& \underset{\left \| \tilde{x} \right \|\leq 1}{\text{sup}}\; a^T_j(\textbf{C}\tilde{x}+\textbf{d})\leq b_j,\,\forall j = 1,\cdots,N
\end{aligned}
\end{equation}
where $a_j$ is a vector representing the $j^{th}$ row of  $\textbf{A}$, $b_j$ is the $j^{th}$ element of $\textbf{b}$, $N$ is the number of obstacles and $K$ is the number of vertices on obstacle $\zeta_j$. The first constraint requires all vertices of obstacles to be on one side of the obtained hyperplanes while the second constraint requires all  points on the ellipsoid to be on the other side of those obtained hyperplanes. 

%

\begin{figure} \vspace{-6pt}
  \centering
  \subfigure[]
  	{\label{fig:unit_axono_view}\includegraphics[width=1.6in]{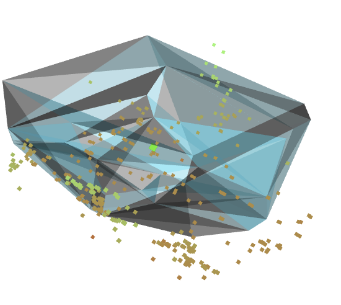}}
  \subfigure[]
  	{\label{fig:unit_top_view}\includegraphics[width=1.6in]{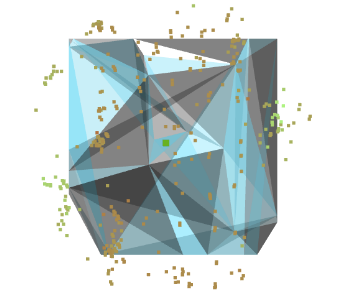}}
  \caption{\small Illustration of a 3D convex hull generated near a seed point and its corresponding nearest neighbors. 
  (a) and (b): The side view and top view of the convex hull. 
  The colorful points are from the seed point's nearest neighbor set $\mathcal{P}_i$.
  } \vspace{-10pt}
\label{fig:3d_convex_hull}  
\end{figure}

\begin{figure*} [t] \vspace{-5pt}
  \centering
  \includegraphics[width=4.5in]{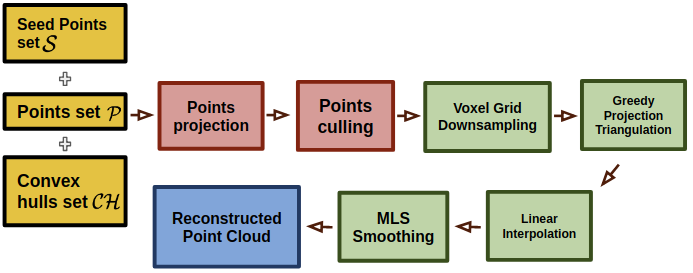}
  \caption{\small The point cloud regulation and refinement flowchart. Yellow blocks are the inputs. Red blocks are the critical steps for noisy point regulation. The green blocks denote the refinement steps, and the blue block is the output of the whole process. }
\label{fig:pcl_procedure} 
\end{figure*}

\subsection{Seeding and Building Local Convex Regions}

A hurdle that prevents us from directly utilizing IRIS lies in that, the obstacles in our case are not convex polyhedrons,  but a cloud of sparse, noisy, and cluttered map points. 

To construct local convex regions, 
first, we sample points as {\em seed points} along the robot's trajectory with a certain interval. Then the evident outlier (noise) map points are filtered and eliminated.
An outlier map point can be identified as a noise if the map point is far away from both the sampling point on robot's trajectory and the centroid of the point cloud around that sampling point. 
To optimize the free space construction,  
instead of searching all obstacle map points to find the points closest to the ellipsoid as mentioned in the standard IRIS process, 
we opt to look for the points closest to the ellipsoid only among a set of nearest neighboring points. 
Namely, for each sampling point, we employ the $k$-nearest neighbors ($k$-NN) method in the filtered point cloud. This step reduces computational load and saves time for remaining operations.

 After that, the convex region growing process starts from each seed point, and terminates when the ellipsoid touches some points or its growth rate is below some predefined threshold. Because such growing process only occurs inside the free space, the final extracted result is always guaranteed to be obstacle-free (and collision-free if the kinodynamic motion is planned inside this space).

In greater detail, let $\mathcal{S} = \{s_i\}$ be the set of seed points sampled from the trajectory, 
and $\mathcal{P}_i$ be the set of nearest neighbors of $s_i$, where $i$ is defined as $i = 1,\cdots, Q$ to index the sampled points along the trajectory.
Each map point from the point cloud is represented as $p_{i}^m$ where $m$ denotes the point index in set $\mathcal{P}_i$. 
The set of convex hulls is represented by $\mathcal{CH}$. 
From each convex hull $ch_i \in \mathcal{CH}$, we can obtain a set of vertices $\mathcal{V}_i$ and a set of meshes $\mathcal{M}_i$.
To create the free space, 
we apply IRIS on each seed $s_i\in \mathcal{S}$ and its  $\mathcal{P}_i$. 
The free space is thus represented with a sequence of convex hulls that are pair-wise overlapped (connected). 
An illustration of a convex hull computed around one seed point is shown in Fig.~\ref{fig:3d_convex_hull}. 
(The piece-wise convexity allows convex constraints to be naturally added while generating robot trajectories, after which the convex optimization problem can be conveniently  formulated and solved.) 

  

\begin{figure} 
  \centering
  \subfigure[]
  	{\label{fig:projection}\includegraphics[width=1.7in]{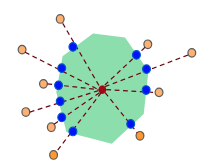}}
  	\ \ 
  \subfigure[]
  	{\label{fig:culling_1}\includegraphics[width=1.5in]{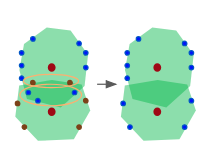}}
  \caption{\small (a) The projection operation. The red point is the seed point $s_i$, the orange points denote $s_i$'s nearest neighbor set $\mathcal{P}_i$, and the blue points denote the projection point set   $\tilde{\mathcal{P}_i}$. The dashed lines represent the visibility lines and the green region is the free space. (b) The culling operation. Pairwise convex hulls have overlapped volumes. This will cause some points (circled) to fall inside two convex hulls.}
\label{fig:post_processing}  
\end{figure}

\subsection{Point Cloud Regulation}


We are also interested in generating the form of point cloud that is similar to high-definition 3D ranging sensors because a dense and clean form of point cloud can help human quickly recognize and comprehend  environmental information. Such ``cleaned" map point cloud is particularly useful for the human-in-the-loop system. 
To achieve this, we propose a method that regulates the original noisy map points by projecting all points on the corresponding convex hull surfaces.

The high-level flowchart can be seen in Fig.~\ref{fig:pcl_procedure}. The input for the whole reconstruction process consists of the set of seed points sampled from the trajectory,  their corresponding  nearest points, and the generated convex hulls using IRIS. 
For each seed point, 
we first make a projection of its nearest points onto the adjacent surfaces of the convex hull. 
Next, since the convex hulls are pairwise overlapped, 
the projection of nearest points on hull surfaces may cause some points to fall inside other convex hulls, and thus those points have to be removed. 
After that, the points will be down-sampled using the {Voxel Grid Filter} approach~\cite{rusu2011point}. 
This allows us to obtain points that are uniformly distributed. 
The down-sampled points will then be triangulated using projection-based incremental triangulation method~\cite{marton2009fast} to form a mesh and by doing so, we create mesh edges between points. 
After building the mesh, we perform an interpolation and add more points on the edges of the built mesh, resulting a denser point cloud. 
Finally, we use {Moving Least Squares}~(MLS)~\cite{scheidegger2005triangulating} to smooth the points, and the output is the reconstructed point cloud. 
In the flowchart (Fig.~\ref{fig:pcl_procedure}), the blocks in green {\em refine} the points and can be achieved using existing tools (e.g., using suites from PCL~\cite{pcl,rusu2011point}). 
In our work we focus on describing the point {\em regulation} in red blocks. More details are as follows.

\subsubsection{Points projection}
For each convex hull $ch_i$, we project its sampling point's nearest neighboring points in $\mathcal{P}_i$ on the meshes $\mathcal{M}_i$ and obtain a set of projected points $\tilde{\mathcal{P}_i}$. See Fig.~\ref{fig:post_processing}. 
Let $\tilde{\mathcal{PC}_i}$ be the points that fall inside other convex hulls in $\tilde{\mathcal{P}_i}$ and let  $\tilde{\mathcal{PC}_i^C}$ be the points that do not fall inside other convex hulls in $\tilde{\mathcal{P}_i}$, i.e., $\tilde{\mathcal{PC}_i^C} = \tilde{\mathcal{P}_i}\backslash \tilde{\mathcal{PC}_i}$. 
This point projection process ensures that the regulated point cloud captures the general structure of the convex hull $ch_i$ while respecting the {\em density distribution} of the original map points. 


\begin{figure} 
  \centering
  \subfigure[]
  	{\label{fig:culling_2}\includegraphics[width=1.6in]{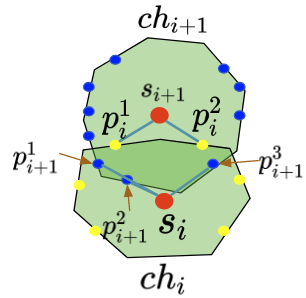}}
  	\ \ 
  \subfigure[]
  	{\label{fig:culling_3}\includegraphics[width=1.6in]{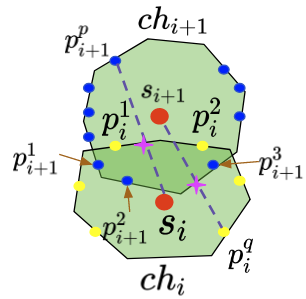}}
  \caption{\small The culling operation is performed for two adjacent convex hulls. We judge whether the points need to be removed by using the visibility lines. A visibility line is the segment connecting a seed point and a projected point. If the visibility segment between a seed point and a projected point is inside a convex hull, this projected point is determined as the point to be removed. }
\label{fig:culling}\vspace{-10pt}
\end{figure}

\subsubsection{Points culling}
The culling operation is performed for two adjacent convex hulls.  Our method for points culling is based on the convexity property of the convex hull, i.e., {a segment connecting any two points inside a convex hull has no intersection with the hull boundary.} 
In our case, if any projected point on a convex hull A falls inside another hull B, the segment that connects this projected point with the seed point of B will be fully inside B. 
On the contrary, if the segment has intersection with the boundary of B, that means the projected point of A is outside the hull B.  

We judge whether the points need to be removed by using  {\em visibility lines}. A visibility line is the segment connecting a seed point and a projected point. If the visibility segment between a seed point and a projected point is fully inside a convex hull, this projected point is deemed as a point lying on the overlapping parts of two convex hulls, and should be culled as it is no longer on the hull surface.

An illustration in 2D scenario is shown in Fig.~\ref{fig:culling}. We take two convex hulls: $ch_i$ and $ch_{i+1}$. The red dots $s_i$ and $s_{i+1}$ are the two adjacent seed points. The projected points $\mathcal{\tilde{P}}_i$ on $ch_i$ are those yellow points in Fig.~\ref{fig:culling_2} whereas the projected points $\mathcal{\tilde{P}}_{i+1}$ on $ch_{i+1}$ are those blue points. The points in $\mathcal{\tilde{PC}}_i$ are notated with $p_i^1$ and $p_i^2$, and the points in  $\mathcal{\tilde{PC}}_{i+1}$ are marked with $p_{i+1}^1$, $p_{i+1}^2$ and $p_{i+1}^3$. 
In Fig.~\ref{fig:culling_2}, all segment lines connecting the seed point $s_i$ and the points in $\mathcal{\tilde{PC}}_{i+1}$ are within $ch_i$. Similarly, all the segment lines connecting the seed point $s_{i+1}$ and the points in $\mathcal{\tilde{PC}}_i$ are within $ch_{i+1}$. In contrast, in Fig.~\ref{fig:culling_3}, any segment line connecting the points in $\mathcal{\tilde{PC}}_i^C$ and $s_{i+1}$ has an intersection with the border of $ch_{i+1}$, whereas any segment line connecting the points in $\mathcal{\tilde{PC}}_{i+1}^C$ and $s_{i}$ causes an intersection with the border of $ch_{i}$. See the purple star in Fig.~\ref{fig:culling_3}.

After point culling, the obtained map points will be located on the outer surface of the convex hulls. 
This culling process allows us to harvest those ``safe'' map points that can contribute to the ultimate navigable free space construction.   
Note however, the resultant map points might not be sufficiently smooth and dense. 
As discussed earlier, refinement techniques shown in green blocks of Fig.~\ref{fig:pcl_procedure} can then be employed to further refine the result.

\section{Experiment}
\label{sect:experiment}

We validate our proposed method using both KITTI dataset and a real built structure/environment.
The sparse map points are generated through running the ORB\_SLAM2~\cite{mur2017orb}. 
For the real environment experiment, we use TurtleBot3 Waffle which is equipped with Intel RealSense R200 camera
carried by the robot for capturing the environment data. 
We use the monocular mode of ORB\_SLAM2 so that only 2D images are used for extracting visual features and a cloud of sparse and noisy map points can be collected. 

\begin{figure} [t] 
  \centering
  \subfigure[]
  	{\label{fig:before_filter_1}\includegraphics[width=1.6in]{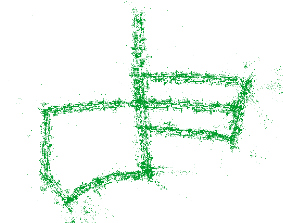}}
  	\subfigure[]
   	{\label{fig:after_filter_1}\includegraphics[width=1.6in]{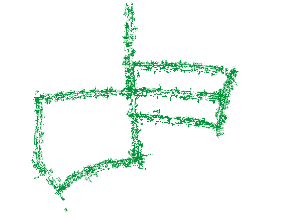}}
   \caption{\small (a) The raw point cloud obtained directly from ORB\_SLAM2; (b) The filtered point cloud using the {Statistical Outlier Removal Filter} method. }
 \label{fig:comparison}  
 \end{figure}

\subsection{Experiment with KITTI Dataset}
\begin{figure} [t] 
  \centering
  \subfigure[]
  	{\label{fig:kitti05_1}\includegraphics[width=3in]{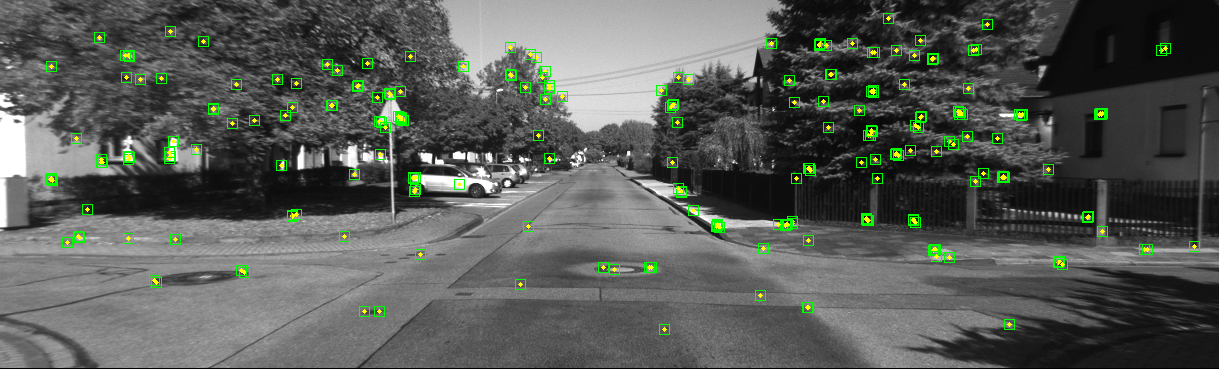}}
   	\subfigure[]
   	{\label{fig:iris_result}\includegraphics[width=3.2in]{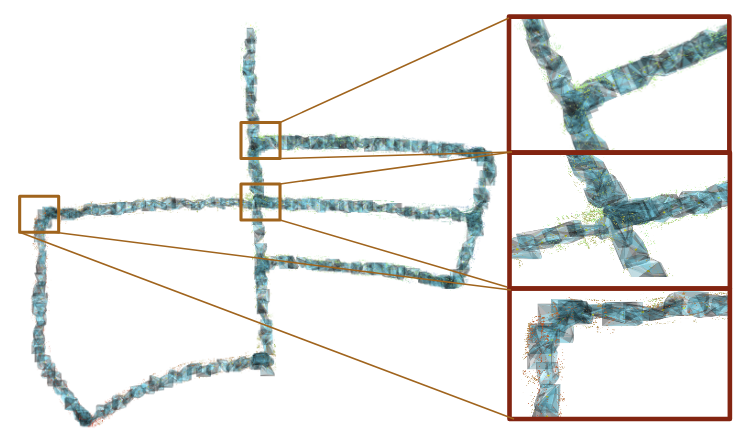}}
   \caption{\small (a) A scene captured in KITTI05 dataset. (b) The created free space consists of a series of overlapping convex hulls. The facets of the convex hulls are represented by blue meshes while the points around the hulls are the nearest neighboring points. We treat these points as obstacles during the convex hull growing process. }
 \label{fig:comparison}  
 \end{figure}

As mentioned in Sect.~\ref{sect:method},  outlier points from the original point cloud need to be eliminated first. 
To do so, the {Statistical Outlier Removal Filter} tool provided by PCL~\cite{pcl} is adopted. 
A comparison between before and after filtering can be seen in Fig.~\ref{fig:comparison}. 
The filtered result provides a point cloud which is fed to our framework.
The created free space using our proposed method is shown in Fig. \ref{fig:iris_result}. 
The convex regions can well cover the free space inside the point cloud and have no obstacle point included. 

\begin{figure} [t]
  \centering
  \subfigure[]
  	{\label{fig:before_2}\includegraphics[width=1.6in]{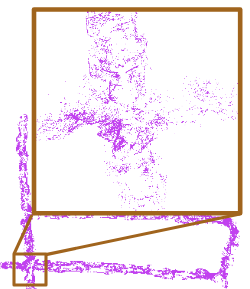}}
  	\ \ 
  \subfigure[]
  	{\label{fig:after_2}\includegraphics[width=1.6in]{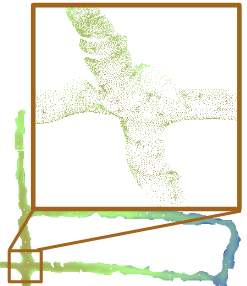}}
  \caption{\small (a) The regulated points on the surfaces of convex hulls. (b) The refined point cloud. 
  }
\label{fig:point_comparison}  
\end{figure}

Based on the constructed free space, we implement the procedure shown in Fig.~\ref{fig:pcl_procedure} and obtained the results as shown in Fig.~\ref{fig:point_comparison}. 
Fig.~\ref{fig:before_2} reveals the regulated point cloud after point projection on convex hulls, from which we can observe that the cloud is cleaner than the original one. 
We then further refine the result and obtain a denser and smoother point cloud as shown in Fig. \ref{fig:after_2}. 
This improved point cloud well shapes the free space while keeping the general point density distribution of the original cloud. 

\subsection{Experiment in Real Environment}
\begin{figure} 
  \centering
       \includegraphics[width=3.4in]{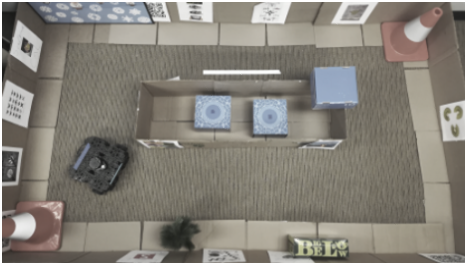}
  \caption{\small To quantitatively evaluate our method with ground-truth information, we built a real environment with dimensions exactly measured. \vspace{-10pt}}
\label{fig:env_13} 
\end{figure}

Although KITTI dataset provides great convenience for conducting simulation, it is extremely difficult to evaluate the performance as the outdoor environment captured in the video is complex and cannot be accessed.
To evaluate a set of algorithmic performances, 
we build an environment with a regular shape whose geometry can be exactly measured and calculated. The built ground-truth environment is shown in Fig.~\ref{fig:env_13}.
We purposely built the environment in a square shape, and its exact dimensions allow us to compare and analyze our reconstruction results in a quantitative way.


\begin{figure} 
  \centering
  \subfigure[]
  	{\label{fig:top_view}\includegraphics[width=1.6in]{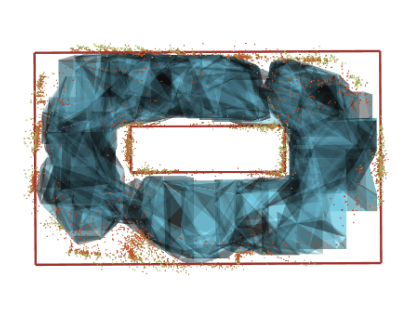} \vspace{-10pt}}
  	\ \ 
  \subfigure[]
  	{\label{fig:other_view}\includegraphics[width=1.6in]{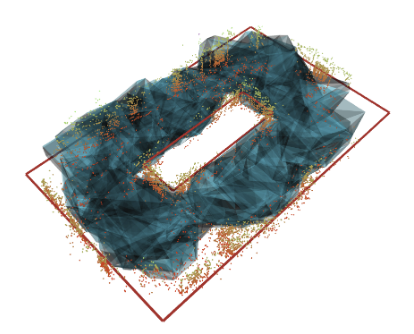}}
  	\ \ 
  \subfigure[]
  	{\label{fig:pc_1}\includegraphics[width=1.6in]{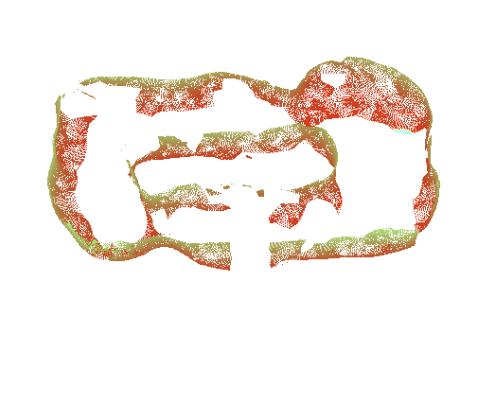}}
  	\ \ 
  \subfigure[]
  	{\label{fig:pc_2}\includegraphics[width=1.6in]{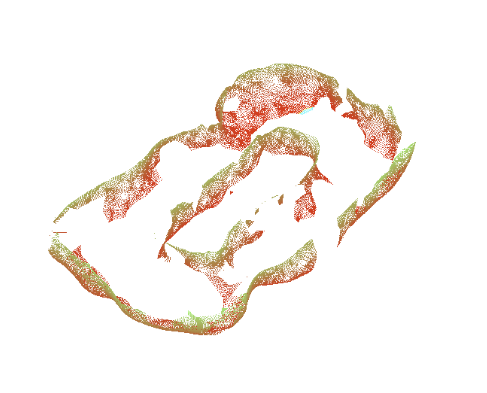} \vspace{-10pt}}
  \caption{\small (a) The top view of the created free space. The red solid line represents the boundary of the obstacles while the dots with varying colors are the point cloud generated from a visual SLAM method. The free space consists of a series of convex hulls. (b) Axonometric view of the created free space. (c) Top view of the reconstructed point cloud. (d) Axonometric view of the reconstructed point cloud.}
\label{fig:view}  \vspace{-10pt}
\end{figure}

To avoid environmental symmetry which could easily provide features of great similarity and result in larger chances of loop closure errors, we design the testing environment in a simple and asymmetric form. 
Additionally, random textures and small objects are placed in the environment to guarantee that
the visual SLAM can detect sufficient features to proceed. 
Our reconstruction result is shown in Fig.~\ref{fig:view}.

\begin{figure} [t] \vspace{-5pt}
  \centering
  	{\label{fig:origin_2}\includegraphics[width=3in]{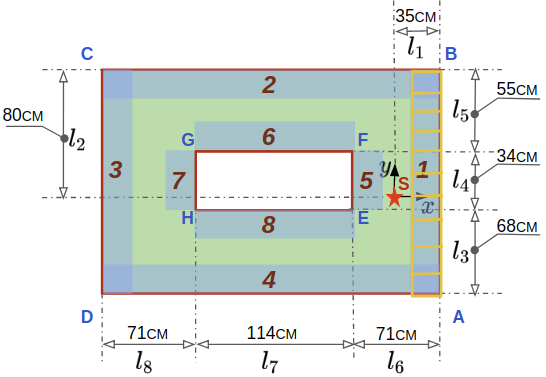}}
  \caption{\small 
  For each boundary a vicinity area is investigated. The areas are labeled with red bold numbers. 
  To discretize the constructed area, a set of 2D grids are used. The yellow grids in area $1$ are shown as an example. The red star represents the origin of the map frame (i.e., the robot's starting position). }
\label{fig:origin}  
\end{figure}

\begin{figure} \vspace{-5pt} 
  \centering
       \includegraphics[width=3.4in]{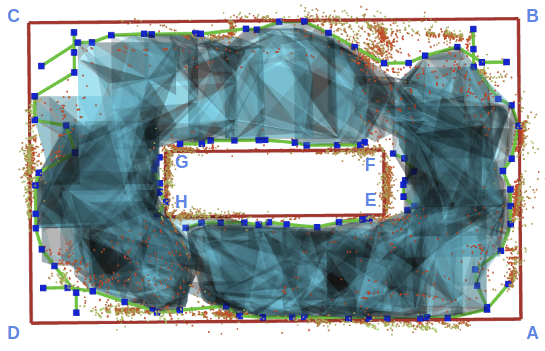}
  \caption{\small Top view of the created free space and the computed approximate boundary points, shown as dark blue points. The green lines represent the approximate boundaries of the free space.}
\label{fig:boundary} \vspace{-10pt}
\end{figure}

To evaluate the experiment result, we compare the boundary of the created free space, that is, the boundary of the convex hulls with the real boundary of the environment. 
The error is defined as the distance between these two versions of boundaries.
Since calculation of the difference between two 3D hull-surfaces is not easy, we thus make some approximations.
First, we project the vertices of the convex hulls onto the $x-y$ plane. Then we separate the boundary of the environment into $8$ segments which form $8$ areas as shown in Fig.~\ref{fig:origin}. 
For each area, we use some cells to discretize it, for example, see the yellow cells in area $1$ in Fig.~\ref{fig:origin}. Some of the previously projected vertices will fall into the cells,
and for each cell we pick the point that is closest to the boundary of the environment as the boundary point of our created free space. For those cells that do not contain any projected points, we take the middle point of the most inner edge of the cell as the boundary point of free space, and we name those middle points as false points. The places which have false points typically have large reconstruction errors, which will be seen later. 
Finally, we obtain a set of approximate boundary points of the created free space. Connecting those points will form an approximate boundary line. 
The distance from an approximate boundary point to the corresponding boundary edge represents the reconstruction error for the corresponding cell. The reconstruction error for each edge could be calculated through the sum of squared differences (SSD). The approximate boundary lines in our experiment result is shown in green lines in Fig.~\ref{fig:boundary}. 

\begin{figure} 
  \centering
       \includegraphics[width=2.6in]{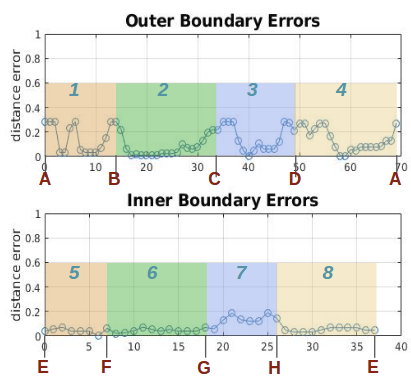}
  \caption{\small Boundary errors are calculated in the respective areas. The areas' labels and the corners of the environment are marked with numbers and capital letters, respectively.}
\label{fig:error} 
\end{figure}

\begin{figure} 
  \centering
  \subfigure[]
  	{\label{fig:number_points_3}\includegraphics[width=2.9in]{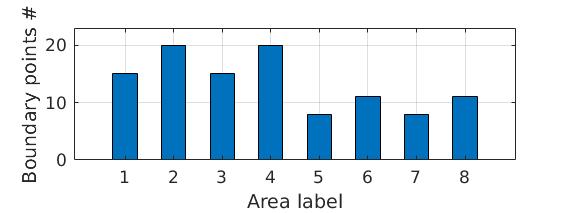}}
  	\ \ 
  \subfigure[]
  	{\label{fig:error_comparison_2}\includegraphics[width=2.9in]{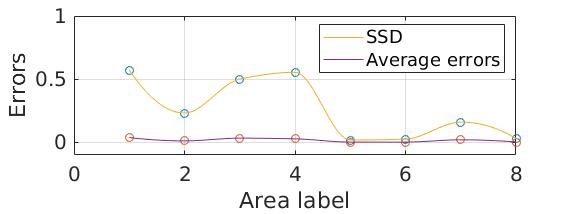}}
  \caption{\small (a) Number of approximated boundary points in each area. (b) For each region, 
  the SSD is calculated. 
  The y-axis denotes the SSD of all hull vertex points (blue points in Fig.~\ref{fig:boundary}) in the monocular SLAM scale/unit (roughly, $1$m $=$ $1.2$unit). 
  The average squared error is the corresponding SSD error divided by the number of boundary points. \vspace{-10pt} 
  } \vspace{-10pt}
\label{fig:table_error}  
\end{figure}

To facilitate evaluation of the reconstruction error, 
we divide the areas into two types: (1) outer areas including areas $1-4$ and (2) inner areas including areas $5-8$, as illustrated in Fig.~\ref{fig:origin}.
Based on the error metric defined above, 
the reconstruction error is shown in Fig.~\ref{fig:error}. 
In our experiment, we measure the unit/scale between the real environment dimension and the ORB\_SLAM2 map dimension, and obtain their scale relation as: $1$m $=$ $1.2$unit. 
The statistics are given in Fig.~\ref{fig:table_error}. 

From Fig.~\ref{fig:error}, we can see that both outer boundary errors and inner boundary errors tend to grow when the robot moves to the corner of the built environment. This is due to the intrinsic limitation of visual SLAM methods  where visual feature points are harder to be detected while the camera is rotating, causing the number of feature points to decrease and the estimation error to increase around corners.
Comparing between the outer boundary errors and inner boundary errors, we can also observe that the values of the former are larger than the later, which also could be seen in Fig.~\ref{fig:error_comparison_2}. 
We can see that both two errors are very small when the robot moves in a straight path along a corridor. 
Such varying property of errors is consistent with the characteristics of visual SLAM methods which work particularly well for straightline movements but can be less robust for rotations.

\begin{figure} \vspace{-5pt} 
  \centering
       \includegraphics[width=0.8\linewidth]{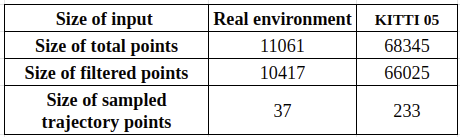}  \vspace{5pt}
  \caption{The size of inputs to our proposed method in real environment experiment and with KITTI dataset.}
\label{fig:input_table} \vspace{-10pt}
\end{figure}

\begin{figure} \vspace{-5pt} 
  \centering
       \includegraphics[width=1\linewidth]{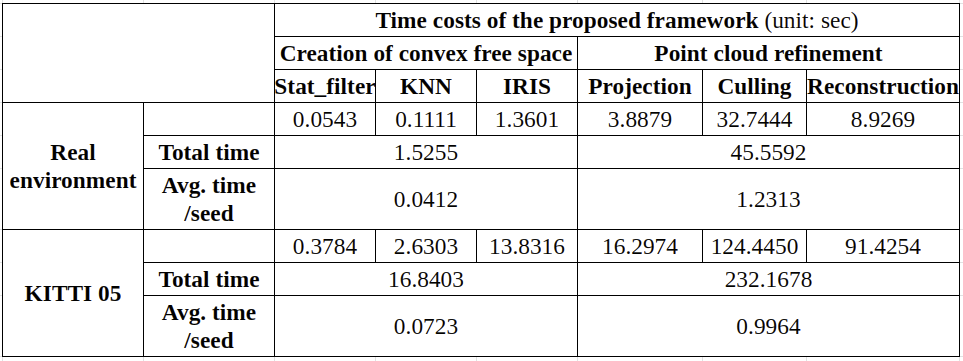} \vspace{5pt}
  \caption{Time costs for different parts of proposed method in real environment experiment and with KITTI dataset.}
\label{fig:timing_table} \vspace{-10pt}
\end{figure}

The time costs of different parts in our proposed framework are presented in Table~\ref{fig:timing_table}, where the  input point clouds information is given in Table~\ref{fig:input_table}. 
As seen in Fig.~\ref{fig:timing_table}, the total time for creating the free space for the {whole map} in our real environment experiment is only around 1.5 seconds whereas the KITTI 05 sequence requires 16.8403 seconds.
However, if we examine the time consumed per seed point (a seed is sampled around every one second), they are in the same level, i.e., the times consumed per seed in real environment and with KITTI 05 sequence are 0.0412 second and 0.0723 second, respectively.
Such limited time cost completely satisfies the requirement for real-time planning and navigation within finite horizons.

For the part of dense point cloud reconstruction, it is relatively time-consuming as revealed in Fig.~\ref{fig:timing_table}. 
However, the regulated and refined dense point cloud is not required for robot navigation and planning: 
the dense point cloud facilitates human recognition and understanding of the environment,  but this component only needs to be executed periodically or based on the human's infrequent requests instead of real-time updates. 
Robot navigation and planning will only require convex hull constructions which can be easily achieved with  real-time computation, as analyzed above.

\section{Conclusion and Future Work}

In this paper, we have presented a framework to create navigable space from sparse and noisy map point cloud generated by existing odometry or mapping methods. Our method first builds local convex regions, from which we further regulate and refine the original noisy and sparse map points to obtain a denser and smoother point cloud that well describes the environment.
We have validated our proposed framework using both a public dataset and a real environmental structure, and our proposed method is validated to be robust to highly noisy map points and efficient for real-time planning and navigation.

{
\bibliographystyle{IEEEtran}

\bibliographystyle{unsrt}
\bibliography{reference}

}

\end{document}